\documentclass[runningheads]{llncs}
\usepackage{marvosym}
\usepackage{graphicx}
\begin{document}
\title{Multi-view Subspace Adaptive Learning via Autoencoder and Attention}
%
%
\author{Jian-wei Liu\inst{1}\textsuperscript{\Letter}\and
Hao-jie Xie\inst{1}\and
Run-kun Lu\inst{1}\and
Xiong-lin Luo\inst{1}}
\authorrunning{Jian-wei Liu et al.}
%
\institute{Department of Automation, College of Information Science and Engineering,
\\China University of Petroleum, Beijing Campus (CUP), 102249, China
\email{liujw@cup.edu.cn}\\
}
\maketitle              
\begin{abstract}
Multi-view learning can cover all features of data samples more comprehensively, so multi-view learning has attracted widespread attention. Traditional subspace clustering methods, such as sparse subspace clustering (SSC) and low-ranking subspace clustering (LRSC), cluster the affinity matrix for a single view, thus ignoring the problem of fusion between views. In our article, we propose a new Multi-view Subspace Adaptive Learning based on Attention and Autoencoder (MSALAA). This method combines a deep autoencoder and a method for aligning the self-representations of various views in Multi-view Low-Rank Sparse Subspace Clustering (MLRSSC), which can not only increase the capability to non-linearity fitting, but also can meets the principles of consistency and complementarity of multi-view learning. We empirically observe significant improvement over existing baseline methods on six real-life datasets.

 \keywords{Multi-view learning \and Subspace self-representation \and Autoencoder \and Attention \and Spectral Clustering.}
\end{abstract}
\section{Introduction}
In real-world machine learning problems, the same data consists of several different representations or views. For example, a traditional web page contains a lot of information. We can use pictures as one view and text features as another view. Therefore, each view reflects some properties of object. Although we can utilize a single view for learning tasks, integrating supplementary information from different views can reduce the complexity for a given task \cite{ref_article1}. In recent years, multi-view learning has attracted more attention. Multi-view learning methods can learn each view with the help of consistency and complementarity on multiple views. Therefore, Multi-view learning not only can effectively use the special information of each view, but also take advantage of the common information of multiple views.

In recent years, many multi-view learning algorithms have been developed. For example, non-negative matrix factorization (NMF) \cite{ref_article2} based multi-view learning algorithms \cite{ref_article3,ref_article4,ref_article5,ref_article6,ref_article7}. These multi-view learning methods all consider consistency and complementarity in multi-view. However, the subspace clustering algorithm often ignores this information. 

The subspace clustering algorithm is commonly divided into two stages. First, we need to construct an affinity matrix for each pair of data points. Second, we use this affinity matrix to implement spectral clustering. Of course, up to now self-representation subspace clustering algorithm is the most representative one. Self-representation-based methods represent data points as a linear combination of other points in the same subspace. These approaches are becoming increasing popular due to their excellent performance. Most recent work on self-representation-based method has focused on incorporating regularization term to make the self-representation matrix more robust, which is better than the factorization method, and can make full use of all data points to obtain a better representation. In this decade, several attempts have been made in this direction. The typical variants are Sparse Subspace Clustering (SSC) \cite{ref_article8}, Low Rank Subspace Clustering (LRSC) \cite{ref_article9}, Consistent and Specific Multi-view Subspace Clustering (CSMSC) \cite{ref_article10}, Multi-view Low-Rank Sparse Subspace Clustering (MLRSSC) \cite{ref_article11}. Among them, CSMSC realizes the principle of consistency and complementarity by decomposing the self-representing coefficient matrix, and MLRSSC explores the alignment of the self-representing coefficient matrix. In addition, there are many methods for subspace self-representation learning using deep neural networks, such as Generalized Latent Multi-view Subspace Clustering (gLMSC) \cite{ref_article12}, Deep Subspace Clustering Networks (DSCN) \cite{ref_article13}.

In this paper, inspired by the autoencoder \cite{ref_article16}, attention mechanism \cite{ref_article17} and MLRSSC, we develop a new Multi-view Subspace Adaptive Learning based on Attention and Autoencoder (MSALAA). First, we map different views to the same dimension, fuse each view with other views through the attention mechanism, and then construct the self-representation. Finally the self-representation output of each view is input to the decoder to reconstruct the original data, which are trained according to our designed loss function. Since the traditional subspace clustering algorithm needs to use ADMM algorithm to iteratively update the training parameters, and our algorithm is implemented by neural network, we only need to choose the optimization strategy, such as SGD, Adam, RMSProp, etc., and the deep learning optimizer will can automatically help us to update parameters. Our contributions in this article are as follows:

(1) We incorporate autoencoder with attention mechanism. The entire deep network is divided into four consecutive parts: encoder layer, multi-view attention layer, self-representation layer, and decoder Layer. The off-the-shelf optimizer of deep learning framework is used to automatically derive and update network parameters.

(2) Inspired by MLRSSC, we fuse the encoders’ output with the same dimensions, so that learning hidden representation for each view incorporates the characteristics of hidden representation of other views. We improve the effect of the self-representation matrix of each view by adapting each view with other views, and explicitly carry out the consistency and complementarity in multi-view learning.

(3) We have conducted extensive experiments on six real-life datasets that have different properties and scales to demonstrate the effectiveness and efficiency of our proposed formulation.
\section{Based Method}
Before introducing our method, we will introduce some basic concepts and representative methods of multi-view subspace clustering.
\subsection{Self-representation of Data}
We briefly introduce the self-representation method for training data. The meaning of data self-representation is that each data point in a union set of subspace can be effectively reconstructed by combining other points in the data set. More precisely, each data point ${{x}_{i}}$ can be expressed as:   
\begin{equation}
x_{i}=\textbf{\textup{X}}c_{i}, c_{ii}=0	
\end{equation}                     
where ${{c}_{i}}=[{{c}_{i1}},{{c}_{i2}},...,{{c}_{iN}}]$ and $N$ represents the number of samples. In addition, ${{c}_{ii}}=0$ represents a simple way of eliminating the linear combination of writing points as themselves. In this way, we can represent each data point in the data point matrix $X$ as a linear combination of other data points.
Since the above problem has solution vectors of infinite number, incorporating constraint, the problem (1) is transformed into the following minimizing problem:
\begin{equation}
	\min {{\left\| {{c}_{i}} \right\|}_{q}}\quad s.t.\quad {{x}_{i}}=\textbf{\textup{X}}{{c}_{i}},\quad {{c}_{ii}}=0 
\end{equation}              
where different choices for$q$ have different effects in the obtained solutions, such as  ${{L}_{1}}$ norm, kernel norm, and F-norm, etc. $q=1$ is used in the SSC algorithm.
We can also rewrite the above problem in matrix form:
\begin{equation}
	\min {{\left\| \textbf{\textup{C}} \right\|}_{q}}\quad s.t.\quad \textbf{\textup{X}}=\textbf{\textup{X}}\textbf{\textup{C}},\quad diag(\textbf{\textup{C}})=0  
\end{equation}           
where $\textbf{\textup{C}}=[c_{1},c_{2},\cdots,c_{N}]\in R^{N\times N} $ is a matrix of self-representation coefficients.
\subsection{MLRSSC}
Our work are inspired by MLRSSC, before we dive into the details of our proposed framework, let's briefly introduce MLRSSC based on pairwise similarity. Because this one is intendedly designed for multi-view data, it has good reference value.
MLRSSC mainly concerns the similarity between matrix pairs represented by self-representation matrices. MLRSSC solves the following joint optimization problems with ${{n}_{\nu }}$ views:
\begin{equation}
\min \limits_{C^{(1)},C^{(2)},\cdots,C^{({n}_{\nu })}}\sum\limits_{v=1}^{n_{\nu }}(\beta_{1}\Vert C^{(v)}\Vert_{*}+\beta_{2}\Vert C^{(v)}\Vert_{1})
+\sum\limits_{1\leq v, w\leq n_{v}, v\neq w}\lambda_{(v)}\Vert C^{(v)}-C^{(w)}\Vert_{F}^2\\
\end{equation}
\begin{equation}
s.t. X^{(v)}=X^{(v)}C^{(v)},diag(C^{(v)}),v=1,\cdots,n_{v} 
\end{equation}
where $C^{(v)}\in R^{N\times N}$ is the self-representation matrix of the view $v$. ${{\beta }_{1}}$,${{\beta }_{2}}$, and ${{\lambda }^{(v)}}$ indicate the trade-off parameters between low-rank, sparse, and consistency constraints between views, respectively. In order to solve the convex optimization problem, MLRSSC used the alternating direction multiplier method (ADMM) \cite{ref_article14}.
\section{Our Proposed framework}
In this section we exposure our proposed multi-view subspace adaptive learning paradigm in which we incorporate autoencoder with attention mechanism. The network structure in MSALAA is shown in Fig. 1. In Fig. 1, for simplicity, we only take the  $v$-th sample with 3 views as a demonstration. 
\begin{figure}
	\centering
	\includegraphics[width=\textwidth]{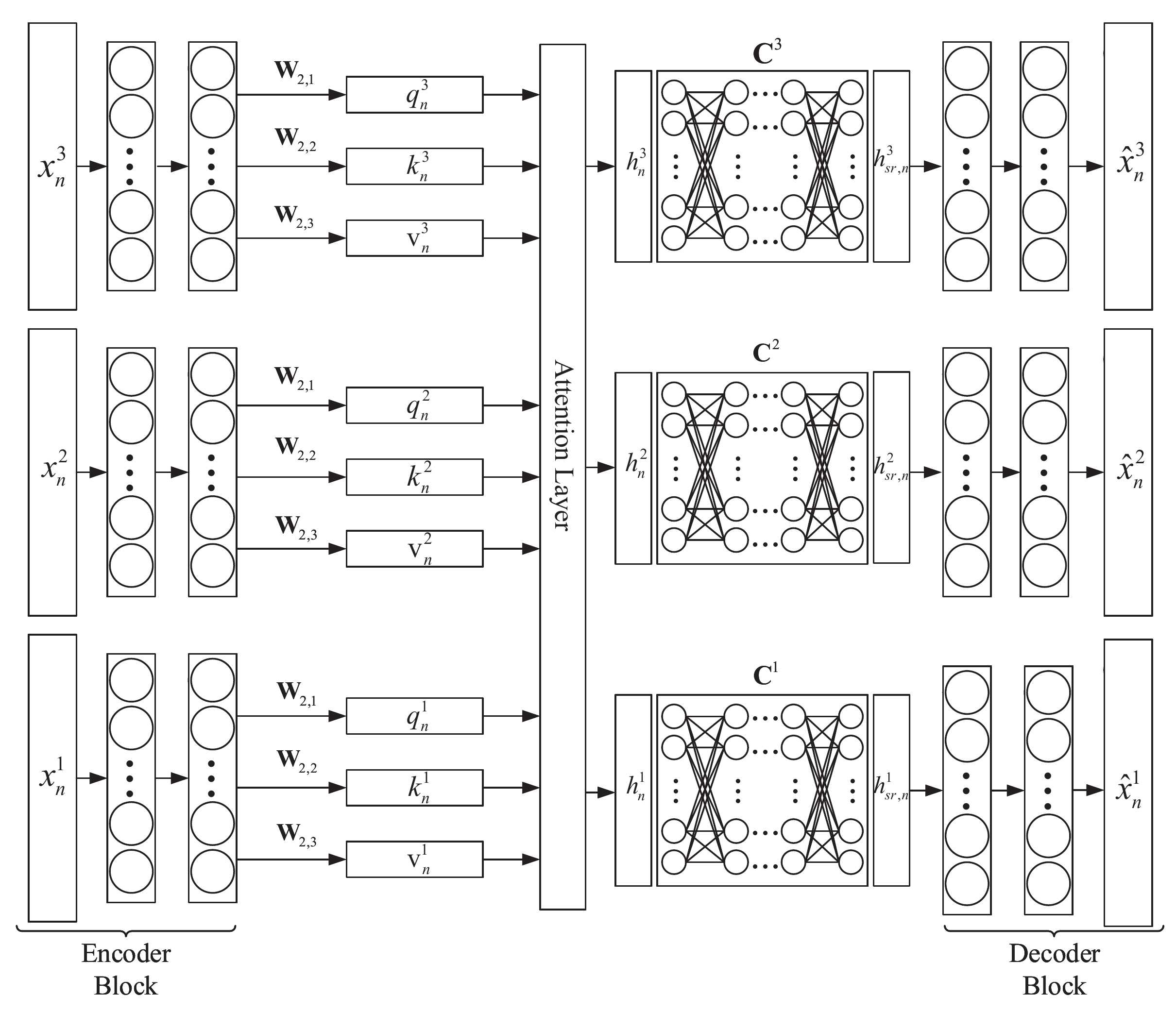}
	\caption{ Network structure of MSALAA.} \label{fig1}
\end{figure}
The entire devised framework is divided into four consecutive stages: encoder layer, multi-view attention layer, self-representation layer, and decoder Layer.
First, let us briefly explain some denotation. Suppose that training data set contains multiple views, the $v$-th view is expressed as follows:
\begin{equation}
X^{v}=[x_{1}^{v},x_{2}^{v},\cdots, x_{n}^{v},\cdots,x_{N}^{v}], X^{v}\in R^{F^{v}\times N},v\in \left\{ 1,\cdots ,M \right\},n\in \left\{ 1,\cdots ,N \right\}
\end{equation}     
where $v$ is the number of $v$-th view, and $N$ is the number of samples. ${{F}_{v}}$ represents the number of features for the $v$-th view.
\subsection{Encoder Layer}
In the training data set, the feature dimensions in different views are different. hence we need to map examples in different views into the same dimension, so that the examples for each view has the same dimension$R$:
\begin{equation}
z_{n}^{v}=f(W_{1,l}^{v}\cdots f(W_{1,2}^{v}f(W_{1,1}^{v}x_{n}^{v}+{{b}_{1,1}})+{{b}_{1,2}})\cdots +{{b}_{1,l}}) 
\end{equation}       

where${{Z}^{v}}=[z_{1}^{v},z_{2}^{v},\cdots ,z_{n}^{v},\cdots ,z_{N}^{v}]\in {{{R}}^{R\times N}} $. $W_{1,1}^{v}\in {{{R}}^{R\times {{F}^{v}}}},\cdots ,W_{1,l}^{v}\in {{{R}}^{R\times {{F}^{v}}}}$and ${{b}_{1,1}}\in {{{R}}^{R}},\cdots,{{b}_{1,l}}\in {{{R}}^{R}}$ are the weight matrices and biases of the fully connected layer, $l$ is the number of network layers, and $f(\cdot )$ is the activation function of the fully connected layer. Here we use the ReLu as activation function.

\subsection{Multi-view Attention Layer}
In this layer, we need to use attention mechanism to process each view in order to achieve the fusion of the content of multiple views for each view.
First, we construct a query matrix $Q$, a key matrix $K$, and a value matrix $V$ by following identities:
\begin{equation}
{{Q}^{v}} = W_{2,1}{{Z}^{v}},{{K}^{v}}= W_{2,2}{{Z}^{v}},{{V}^{v}}= W_{2,3}{Z}^{v} 
\end{equation}            
where${{Q}^{v}}=[q_{1}^{v},\cdots ,q_{n}^{v},\cdots ,q_{N}^{v}], {{K}^{v}}=[k_{1}^{v},\cdots ,k_{n}^{v},\cdots ,k_{N}^{v}], {{V}_{v}}=[v_{1}^{v},\cdots ,v_{n}^{v},\cdots ,\\v_{N}^{v}], q_{n}^{v},k_{n}^{v},v_{n}^{v}\in {{{R}}^{R\times 1}}$, and $W_{2,1}^{{}},W_{2,2}^{{}},W_{2,3}^{{}}\in {{{R}}^{R\times R}}$ are the linear transform matrices.
Secondly, we need to calculate alignment weight $a_{i}^{v}$ for the context vector$h_{i}^{v}$ of $i$-th sample in the $v$-th view. We first define the score function:
\begin{equation}
{score(q,k)=q}\cdot k  
\end{equation}                
The alignment weight is defined as follows:
\begin{equation}
a_{i}^{v}=\frac{exp(score(q_i^v,k_i^v))}{\sum_{j=1}^{M}exp(q_i^v,k_i^j)}
\end{equation}                   
By getting the alignment weight $a_{i}^{v}$ we can derive the context vector $h_{i}^{v}$:
\begin{equation}
h_{i}^{v}=a_{i}^{v}v_{i}^{v}
\end{equation}                         
\subsection{Self-representation Layer}
For the subspace clustering, we need to utilize self-representation coefficient matrix ${{C}^{v}}\in {{{R}}^{N\times N}}$ and ${{H}^{v}}=h_{1}^{v},\cdots,h_{N}^{v} \in {{{R}}^{R\times N}}$to obtain representation recombination matrix$H_{sr}^{v}=[h_{sr,1}^{v},h_{sr,2}^{v},\cdots ,h_{sr,n}^{v},\cdots ,h_{sr,N}^{v}]\in {{{R}}^{R\times N}}$, the formula is as follows:
\begin{equation}
H_{sr}^{v}={{H}^{v}}{{C}^{v}}    
\end{equation}                  
What we need to note is that in subspace clustering, the self-representation coefficient matrix ${{C}^{v}}$ needs to satisfy the constraint $diag({{C}^{v}})=0$.
\subsection{Decoder Layer}
In this layer, we utilize the self-representation $h_{sr,n}^{v}$ as input of the fully connected layer to reconstruct the original data $x_{n}^{v}$of each view:
\begin{equation}
\hat{x}_{n}^{v}=f(W_{3,l}^{v}\cdots f(W_{3,2}^{v}f(W_{3,1}^{v}h_{sr,n}^{v}+{{b}_{2,1}})+{{b}_{2,2}})\cdots +{{b}_{2,l}})
\end{equation}       
where $\hat{x}_{n}^{v}$is reconstruction of  $x_{n}^{v}$. In addition, $W_{3,1}^{v}\in {{{R}}^{{{F}^{v}}\times R}},\cdots ,W_{3,1}^{v}\in {{{R}}^{{{F}^{v}}\times R}}$ and ${{b}_{2,1}}\in {{{R}}^{{{F}^{v}}}},\cdots ,{{b}_{2,l}}\in {{{R}}^{{{F}^{v}}}}$ are the weight matrices and bias vectors of the fully connected layer, and $f(\cdot )$ is the activation function of the fully connected layer, which uses the ReLu as activation function.We concatenate $\hat{x}_{n}^{v}$,$n\in \left\{ 1,\cdots ,N \right\}$ to form the matrix ${{\hat{X}}^{v}}=[\hat{x}_{1}^{v},\hat{x}_{2}^{v},\cdots ,\hat{x}_{n}^{v},\cdots,\hat{x}_{N}^{v}]\in {{{R}}^{{{F}^{v}}\times N}}$.

\subsection{Loss Function}
This loss function can be divided into two parts. The first part is related to the encoder and decoder:
\begin{equation}
\sum\limits_{v=1}^{M}{\frac{1}{2NM}\left\| {{X}^{v}}-{{{\hat{X}}}^{v}} \right\|_{F}^{2}+{{\beta }_{2}}\Omega (W_{1,1}^{v},\cdots ,W_{1,l}^{v},W_{3,1}^{v},\cdots ,W_{3,l}^{v})}
\end{equation}
where $\Omega (\cdot )$ is a regularization term, and its role is to constrain the parameters in the encoder and decoder. In this paper, $\Omega (\cdot )$mainly have two forms, one is ${{L}_{1}}$ norm regularization term and the other is ${{L}_{2}}$ norm regularization term, and ${{\beta }_{2}}$ is a trade-off parameter.
The other part is related to the self-representation of the subspace. 
\begin{equation}
\sum\limits_{v=1}^{M}{\frac{1}{2N}\left\| H_{sr}^{v}-{{H}^{v}} \right\|_{F}^{2}+\left\| {{C}^{v}} \right\|_{F}^{2}}+{{\beta }_{1}}\sum\limits_
	{1\le v,w\le M,	v\ne w}  
	{\frac{1}{2N}\left\| {{C}^{v}}-{{C}^{w}} \right\|_{F}^{2}}  
\end{equation}  
Here we first impose the alignment constraint for the self-representation matrices of multiple views in the form of $\sum\limits_{1\le v,w\le M,v\ne w}{\frac{1}{2N}\left\| {{C}^{v}}-{{C}^{w}} \right\|_{F}^{2}}$, by imposing the alignment constraint, the information of each view and other views can be fused with each other, so that the multi-views are complementary. We introduce the F norm for the self-representation coefficient matrix ${{C}^{v}}$. In the subspace self-representation learning task, we can apply different regularization constraints on the self-representation coefficient matrix ${{C}^{v}}$, for example, ${{L}_{1}}$ norm regularization term, kernel norm Constraints, and the F-norm constraints witch we use. In addition, ${{\beta }_{1}}$ is a trade-off parameter.
Due to the use of deep autoencoder architecture, an additional benefit is that we do not need to resort complex optimization approaches, such as the ADMM algorithm for iterative updates, which is used in SSC and LRSC. We can directly update the self-representation coefficient matrix ${{C}^{v}}$ by the off-the-shelf Stochastic Gradient Descent (SGD) approaches in Tensorflow. 

\begin{table}
	\centering
\caption{Characteristics of Six Datasets}\label{tab1}
\begin{tabular}{|l|l|l|l|c|}
\hline
Datasets &  M & C & N & $F^v$\\
\hline
ORL       & 3 & 40 & 400 & [4096,3304,6750] \\
Reuters   & 5 & 6  & 600 & [21526,24892,34121,15487,11539] \\
3-sources & 3 & 6  & 169 & [3560,3631,3068] \\
Yale & 3 & 15 & 165 & [4096,3304,6750] \\
UCI digit & 3 & 10 & 2000 & [216,76,64] \\
Prokaryotic & 3 & 4 & 551 & [438,3,393] \\
\hline
\end{tabular}
\end{table}
\section{Experimental Results}
In this section, we design a series of experiments to demonstrate the effectiveness of MSALAA on real-world data sets. We first introduce the data set we use. Secondly, we describe the baseline methods, and finally we elaborate the configuration of hyper-parameters, and perform spectral clustering with MSALAA and six baseline methods on six data sets.
\subsection{Datasets}
To explore the performance of our proposed network, we have performed comparative experiments on six datasets, which are ORL, Reuters, 3-sources, Yale, UCI-digit, and Prokaryotic dataset. The characteristics of six datasets are summarized in Table 1. Where $M,C,N$ and $F^v$ represents the number of views, the sample category, the number of samples, and the feature dimensions of each view, respectively.
\subsection{Baseline Methods}
To demonstrate the efficiency of our proposed MSALAA, some state-of-the-art subspace clustering methods are chosen as the baseline methods: SSC, LRSC, LMSC \cite{ref_article15}, CSMSC, and MLRSSC and its three improved variants (MLRSSC-Centroid, KMLRSSC, KMLRSSC-Centroid).
\subsection{The Configuration of Hyper-parameters  }
In our experiments, we only need to set the number of layers of the network and the number of neurons in each layer, and determine whether to perform batch normalization. In this article, all our datasets are initialized with the LeCun normal distribution. We choose Adam optimization algorithm as optimizer, the learning rate is 0.001, and the decay value of learning rate is 0.99. In addition, we set the trade-off parameters ${{\beta }_{1}}$ and ${{\beta }_{2}}$ to 0.1. In the experiment, we will get the self-representation matrix of multiple views. Here we will use the self-representation matrix${{C}^{v}}$of the best experimental view to construct the affinity matrix${{A}^{v}}=\left| {{C}^{v}} \right|+{{\left| {{C}^{v}} \right|}^{T}}$for spectral clustering. In addition, we performed residual processing on the network to prevent the gradient from disappearing.
\begin{figure}
	\centering
	\includegraphics[width=0.6\textwidth]{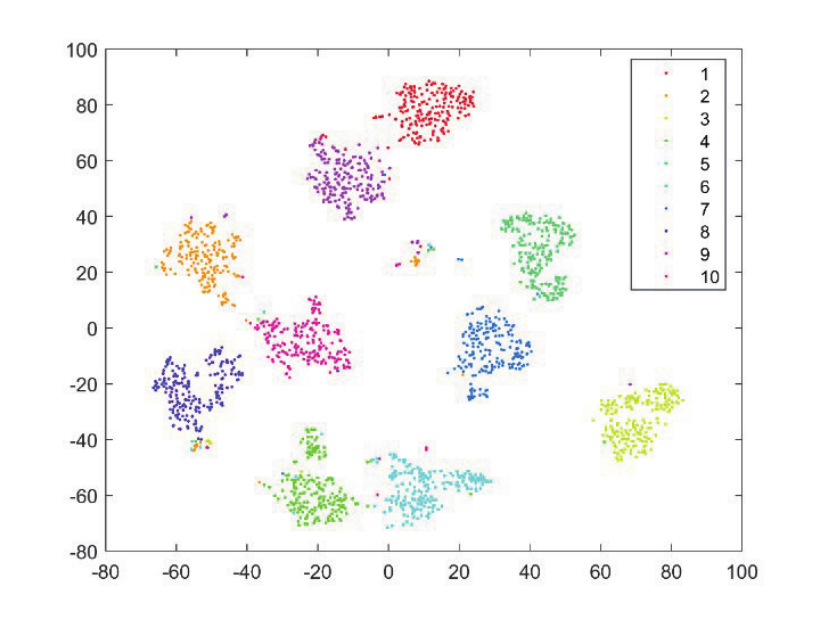}
	\caption{Visualization of ${{H}^{v}}$on UCI-digit.} \label{fig2}
\end{figure}
\subsection{Experimental Results and Analysis}
We perform spectral clustering with MSALAA and six baseline methods on six data sets. In order to verify the performance of our method, we selected six metric criteria to measure the effect of spectral clustering: Accuracy (ACC), Normalized Mutual Information (NMI), Adjusted Rand Index (ARI), Precision, Recall and F-score. From the experimental results in Table 2, we can clearly see that our method has obvious advantages over other baseline comparison methods, validate that MSALAA can find better data self-representation. On the data set UCI-digit, MSALAA can reach more than 90\% on multiple metric criteria.

In addition, we also visualize the matrix ${{H}^{v}}$generated by the attention mechanism. Here we use t-SNE. In the visualization experiment, we use t-SNE to embed feature matrix${{H}^{v}}$into a 2D latent feature matrix for clustering. We use t-SNE to derive a 2D latent feature matrix and depict it with a scatter plot. As shown in Fig. 2, this process is performed on the data set UCI-digit, and we can clearly see that each clustering can be easily distinguished.
\begin{table}
	\centering
	\fontsize{5}{8}\selectfont
	\caption{Experimental Comparison Results on Three Datasets}\label{tab2}
	\begin{tabular}{|c|l|l|l|l|l|l|}
		\hline
		Method for ORL &  ACC & NMI & ARI & Precision & Recall & F-score\\
		\hline
	    SSC &  72.13\%(0.017\%)	& 87.33\%(0.001\%)&63.26\%(0.023\%)&74.25\%(0.023\%)&	72.13\%(0.017\%)&70.84\%(0.022\%)\\
	    LRSC&72.23\%(0.027\%)&84.06\%(0.005\%)&59.82\%(0.032\%)&74.78\%(0.041\%)&	72.23\%(0.027\%)&71.79\%(0.033\%)\\
	    LMSC&82.18\%(0.136\%)&92.74\%(0.018\%)&77.39\%(0.188\%)&81.25\%(0.148\%)&	82.18\%(0.136\%)&	80.21\%(0.168\%)\\
	    CSMSC&84.55\%(0.004\%)&92.74\%(0.001\%)&79.78\%(0.004\%)&	85.30\%(0.006\%)&	84.55\%(0.004\%)&83.78\%(0.005\%)\\
	    MLRSSC&63.68\%(0.116\%)&81.32\%(0.029\%)&52.41\%(0.136\%)&63.97\%(0.108\%)&	63.68\%(0.116\%)&62.27\%(0.113\%)\\
	    MLRSSC-C&78.03\%(0.075\%)&91.65\%(0.008\%)&72.88\%(0.085\%)&78.46\%(0.106\%)&	78.03\%(0.075\%)&76.11\%(0.101\%)\\
	    KMLRSSC&78.55\%(0.166\%)&90.27\%(0.026\%)&72.05\%(0.175\%)&79.32\%(0.163\%)&	78.55\%(0.166\%)&77.43\%(0.177\%)\\
	    KMLRSSC-C&78.25\%(0.094\%)&90.71\%(0.007\%)&72.06\%(0.069\%)&79.34\%(0.153\%)&	78.25\%(0.094\%)&77.16\%(0.123\%)\\
	    MSALAA&86.40\%(0.009\%)&93.16\%(0.009\%)&80.91\%(0.032\%)&86.43\%(0.023\%)&	86.40\%(0.009\%)&85.38\%(0.015\%)\\
	    \hline
	    Method for Reuters &  ACC & NMI & ARI & Precision & Recall & F-score\\
	    \hline
	    SSC&50.85\%(0.000\%)&35.05\%(0.002\%)&24.69\%(0.000\%)&51.93\%(0.001\%)&	50.85\%(0.000\%)&44.89\%(0.000\%)\\
	    LRSC&31.12\%(0.000\%)&14.18\%(0.000\%)&3.257\%(0.000\%)&58.52\%(0.000\%)&	31.12\%(0.000\%)&28.13\%(0.000\%)\\
	    LMSC&41.53\%(0.010\%)&33.04\%(0.024\%)&17.34\%(0.039\%)&40.56\%(0.014\%)&	41.53\%(0.010\%)&33.46\%(0.005\%)\\
	    CSMSC&42.42\%(0.000\%)&32.63\%(0.000\%)&18.28\%(0.000\%)&46.20\%(0.000\%)&	42.42\%(0.000\%)&34.49\%(0.000\%)\\
	    MLRSSC&52.95\%(0.108\%)&38.22\%(0.024\%)&28.18\%(0.070\%)&50.09\%(0.308\%)&	52.95\%(0.108\%)&48.30\%(0.259\%)\\
	    MLRSSC-C&51.35\%(0.124\%)&36.96\%(0.013\%)&26.76\%(0.070\%)&48.36\%(0.160\%)&	51.35\%(0.124\%)&45.89\%(0.216\%)\\
	    KMLRSSC&57.12\%(0.054\%)&37.38\%(0.037\%)&30.41\%(0.040\%)&61.79\%(0.135\%)&	57.12\%(0.054\%)&56.67\%(0.088\%)\\
	    KMLRSSC-C&55.12\%(0.057\%)&35.69\%(0.026\%)&29.38\%(0.027\%)&59.05\%(0.208\%)&	55.12\%(0.057\%)&53.97\%(0.071\%)\\
	    MSALAA&57.88\%(0.034\%)&40.69\%(0.020\%)&31.02\%(0.013\%)&65.14\%(0.025\%)&	57.88\%(0.034\%)&57.00\%(0.043\%)\\
		\hline
		Method for UCI digit &  ACC & NMI & ARI & Precision & Recall & F-score\\
		\hline
		SSC&78.02\%(0.004\%)&79.08\%(0.007\%)&71.06\%(0.011\%)&79.41\%(0.001\%)&	78.02\%(0.004\%)&77.59\%(0.001\%)\\
		LRSC&64.19\%(0.001\%)&68.61\%(0.000\%)&56.01\%(0.003\%)&65.54\%(0.003\%)&	64.19\%(0.001\%)&63.20\%(0.001\%)\\
		LMSC&74.11\%(0.446\%)&74.63\%(0.112\%)&65.19\%(0.274\%)&74.56\%(0.591\%)&	74.11\%(0.446\%)&72.98\%(0.582\%)\\
		CSMSC&83.22\%(0.114\%)&78.48\%(0.017\%)&72.29\%(0.048\%)&83.30\%(0.182\%)&	83.22\%(0.115\%)&82.78\%(0.157\%)\\
		MLRSSC&88.08\%(0.374\%)&85.15\%(0.048\%)&81.37\%(0.251\%)&87.71\%(0.485\%)&	88.08\%(0.374\%)&87.34\%(0.496\%)\\
		MLRSSC-C&89.27\%(0.298\%)&85.32\%(0.054\%)&81.81\%(0.253\%)&89.14\%(0.399\%)&	89.27\%(0.298\%)&88.74\%(0.395\%)\\
		KMLRSSC&89.35\%(0.244\%)&86.08\%(0.033\%)&82.60\%(0.193\%)&88.76\%(0.360\%)&	89.35\%(0.244\%)&88.61\%(0.342\%)\\
		KMLRSSC-C&90.97\%(0.212\%)&86.51\%(0.033\%)&84.03\%(0.155\%)&91.03\%(0.257\%)&	90.97\%(0.212\%)&90.64\%(0.284\%)\\
		MSALAA&96.07\%(0.000\%)&91.31\%(0.001\%)&91.48\%(0.001\%)&96.12\%(0.000\%)&	96.07\%(0.000\%)&96.08\%(0.000\%)\\
		\hline
	\end{tabular}
\end{table}
\section{Conclusion and Future Work}
We propose a new Multi-view Subspace Adaptive Learning based on Attention and Autoencoder (MSALAA) combined with mutual subspace alignment in subspace clustering. Our method takes into account the two important factors of consistency and complementarity in multi-view learning, and also utilizes the neural network to increase the nonlinear representation ability of the model. The experiments on several real-world datasets showed that the proposed MSALAA mostly outperformed the other baseline methods, which validate that our proposed MSALAA can use self-representation of multi-views to subspace adaptive learning.

In future work, we will investigate some variants for MLRSSC. We will try to build a common self-representation matrix ${{C}^{*}}$ and align ${{C}^{*}}$ with all views to see the influence of performance for multi-view subspace Learning. In addition, we can improve the basic fully connected network to achieve better performance on multi-view data.

%
%
%
%

\end{document}